\pdfoutput=1

\documentclass[11pt]{article}

\usepackage[final]{EMNLP2022}

\usepackage{times}
\usepackage{latexsym}

\usepackage[T1]{fontenc}

\usepackage[utf8]{inputenc}

\usepackage{microtype}

\usepackage{inconsolata}

\usepackage{amsmath}
\usepackage{amsfonts}
\usepackage{amsmath}
\usepackage{amsthm}
\usepackage{dsfont}
\usepackage{xcolor}
\usepackage{url}
\usepackage{comment}
\usepackage{natbib}
\usepackage{graphicx}
\usepackage{listings}
\usepackage{hyperref}

\usepackage{multirow}

\newcommand{\carp}[1]{CARP}
\newcommand{\coop}[1]{CoOp}
\newcommand{\pseudo}[1]{Pseudo CARP CoOp}
\newcommand{\alignment}[1]{Alignment CARP CoOp}
\newcommand{\defaultlm}[1]{Default CARP LM}
\newcommand{\pseudolm}[1]{Pseudo CARP LM}
\newcommand{\alignmentlm}[1]{Alignment CARP LM}

\title{Robust Preference Learning for Storytelling 

via Contrastive Reinforcement Learning}

\author{Louis Castricato$^*$ \\
  CarperAI\\
  Brown University \And
  Alexander Havrilla$^*$ \\
  CarperAI\\
  Georgia Tech \And 
  Shahbuland Matiana \\
  CarperAI \\
  University of Waterloo  \\
  \AND
  Michael Pieler\\
  EleutherAI \And 
   Anbang Ye \\
   Georgia Tech \And 
   Ian Yang \\
   Georgia Tech \And 
   Spencer Frazier \\
   Georgia Tech \And 
   Mark Riedl \\
   Georgia Tech
}

\begin{document}
\maketitle
\begin{abstract}
Controlled automated story generation seeks to generate natural language stories satisfying constraints from natural language critiques or \textit{preferences}. Existing methods to control for story preference utilize prompt engineering which is labor intensive and often inconsistent. They may also use logit-manipulation methods which require annotated datasets to exist for the desired attributes. To address these issues, we first train a contrastive bi-encoder model to align stories with corresponding human critiques, named \textit{CARP}, building a general purpose \textit{preference model}. This is subsequently used as a reward function to fine-tune a generative language model via reinforcement learning. However, simply fine-tuning a generative language model with a contrastive reward model does not always reliably result in a story generation system capable of generating stories that meet user preferences. To increase story generation robustness we further fine-tune the contrastive reward model using a prompt-learning technique. A human participant study is then conducted comparing generations from our full system, ablations, and two baselines. We show that the full fine-tuning pipeline results in a story generator preferred over a LLM 20x as large as well as logit-based methods. This motivates the use of contrastive learning for general purpose human preference modeling.
\end{abstract}

\section{Introduction}
\def\thefootnote{*}\footnotetext{These authors contributed equally to this work. Order is alphabetical. Correspondence to \href{mailto:louis_castricato@brown.edu}{louis\_castricato@brown.edu}.}\def\thefootnote{\arabic{footnote}}

Contemporary, controlled automated story generation uses intelligent systems to generate text from a minimal number of inputs---often a simple prompt and some other criteria---that conveys a coherent and consistent sequence of events. These events adhere to a given set of preferences without requiring manual collection of datasets for particular preferences.
Large, pre-trained neural language models possess impressive generation capabilities. These models, however, struggle with coherence and controllability across longer segments of text~\citep{gpt3,yao2019plan}. 

\begin{figure}[t]
    \centering
    \includegraphics[width=\linewidth]{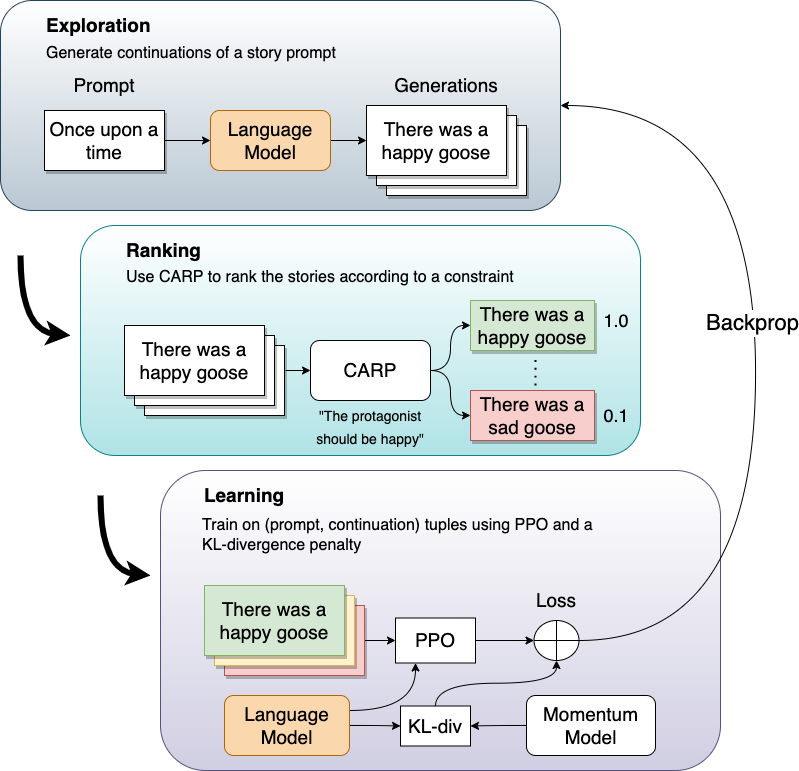}
    \caption{Illustration of our technique for generating story content controlled by preferences.
    A language model generates candidates, which are ranked by the \carp{} model to produce scores. The scores are used to fine-tune the language model to produce higher scoring---and thus more aligned with preferences---story continuations.}
    \label{fig:carppipeline}
\end{figure}

In this paper, we examine whether the consistency of story generation by language models can be improved for a wider class of preferences.
Consider the following: a person wishes to prompt a story generating model to produce a narrative, but also wants the resulting story to hold some subjective property. This could mean the plot of said story ``being sad'' or ``having well formed characters'',  or ``having a sophisticated plot twist."

Conventionally, language models' few-shot prompting capabilities can be prompted with instructions about the subjective quality desired alongside the first line of the story. For some types of preferences, large language models---such as GPT-3 \citep{gpt3} or GPT-NeoX-20B~\citep{Black2022GPTNeoX20BAO}---can consistently meet the criteria. However, for other preferences, such as prompting the moral alignment of a protagonist within a story, performance can be highly variable.

Other methods guide large language models toward particular topics by post-processing the logits. Examples of this include GeDi~\cite{krause2020gedi} and {\em Plug \& Blend}~\cite{lin2021plug}---in both cases a second model then modifies the output logits of the language model and increase frequency of a topic in the narrative.
However logit-based methods - such as GeDi - are often less capable when presented with complex preferences. This in turn limits the model's expressiveness. Further, these methods require potentially expensive collection of annotated datasets aligned to user preference.

To circumvent this explicit data requirement while allowing a large class of preferences we leverage \carp{}~\citep{Matiana2021CutTC}. \carp{} is a contrastively-trained bi-encoder which learns to align a story with a corresponding human critique of the story. Previous work has shown these classifications correlate strongly with a human baseline, demonstrating effectiveness as a general purpose \textit{preference model} for generation in alignment with human preferences.

We use Proximal Policy Optimization (PPO)~\cite{Schulman2017ProximalPO} to fine-tune  GPT-2-750M~\cite{radford2019language} to generate text consistent with a given initial criterion. Reward is represented as the \carp{} similarity of the generated story and the desired preference. Initial attempts indicated the reward signal generated by CARP could sometimes be exploited by the generator, resulting in collapse. In some other cases the generator failed to learn anything at all. 

To address this, we present CARP CoOp: a robust version of CARP leveraging prompt tuning for a stronger reward signal. We deploy a pseudo-labeling technique on CARP's latent space. This allows identification of preferences resistant or susceptible to collapse. Further, we find CARP CoOp is extremely data-efficient, easily incorporating previously unknown preferences with only a couple hundred examples when they are available. This efficiency is demonstrated on a moral alignment dataset which classifies character stories into 'good', 'netural' or 'evil'. We present this as a pipeline for fine-tuning a new language model that can robustly generate text consistent with a complex set of preferences expressed in natural language. 

We evaluate our approach with a human participant study; participants were asked to match sections of generated stories to a list of preference labels ranging from moral alignment to detailed subjective imagery. We show that our proposed technique is better at producing story segments that capture the given preference than prompting a much larger (GPT-NeoX-20B) language model and the GeDi method utilizing logit manipulation. Further, we conduct an ablation study by fine-tuning GPT-2-750M with standard \carp{} but without \coop{}, showing \carp{} can still improve preferences over the NeoX baseline.

In summary, we make the following contributions:
\begin{enumerate}
    \item Introduction of a contrastively trained preference model, \carp{}, as a reward signal for preference learning in story generation.
    \item A new model, \pseudo{}, that improves the robustness of preference learning via \carp{} over a wide class of preferences.
    \item The introduction of the \alignment{} model which signals the moral alignment of story characters. This demonstrates the data efficiency of \carp{} CoOp when annotated data is available.
    \item A human subject study evaluating how well existing and proposed generation methodologies satisfy desired human preferences.
\end{enumerate}

\section{Related Work}

Much of this work is based on prior work related to neural networks---recurrent and transformer-based---capable of producing stories~\cite{roemmele2016writing,khalifa2017deeptingle,martin2018event,clark2018, fan2018hierarchical,ammanabrolu2021automated}.
The \textit{controllability} of neural language models is a central concern in story generation. It remains an open problem as how best to ensure user-desired story properties. Story generation can be controlled by conditioning generation on high-level plot outlines~\cite{fan2018hierarchical,peng2018towards,rashkin2020plotmachines}, story in-filling~\cite{donahue2020ilm,wang2020narrative}, fine-tuning on goals~\cite{tambwekar2019controllable,Alabdulkarim2021GoalDirectedSG}.
Similarly to GeDi, Plug \& Blend~\cite{lin2021plug} learns modifiers to apply to language model logits to control the topic of the generation by training an auxillary topic classification model.

\carp{} is a contrastively-trained bi-encoder trained on paired story text and critique text. 
Like CLIP \citep{Radford2021LearningTV}, a bi-encoder for images and text, it learns to align positive examples of stories adhering to a specific critique and reject negatives.
In the case of \carp{}, a positive example is a story and a critique written expressly for that story. A negative example is a story and any critique of other stories.  \carp{} was trained on the {\em Story Critique} Dataset, composed of 1.3 million story-critique pairs. It has been shown \carp{} matches human preferences more reliably than a baseline auto-regressive model fine-tuned to predict critiques given a passage and vice-versa.
Prior work on preference learning for language models using natural language critiques include \cite{scheurer2022training, saunders2022self}. In particular \citet{Krishna2022RankGenIT} shows the robustness of using a contrastive learning based signal to aid in natural language generation.

We use reinforcement learning via PPO \footnote{Reinforcement learning was performed with TRL, \href{https://github.com/lvwerra/trl/}{https://github.com/lvwerra/trl/}.} and the contrastive \carp{} model as a reward signal for generation. Work by \citep{peng-etal-2020-reducing} trained a language model to generate less non-normative language using a normative text classifier~\citep{Frazier2020LearningNF} to modify the loss function. A similar strategy was used to train a language model to generate stories that ended in a given goal~\citep{ijcai2019-829, Alabdulkarim2021GoalDirectedSG}.
%


Finally, we leverage prompt tuning in the construction of our final \carp{} \coop{} pipeline. Existing CLIP literature shows such approaches can make contrastive-based classification models more robust. In particular {\em Context Optimization} (CoOp)~\cite{Zhou2021LearningTP} is a technique originally designed to optimize prompts for multimodal contrastive models such as CLIP. CoOp produces a sequence of embedded terms \texttt{[V$_1$] [V$_2$] ... [V$_M$] [CLASS]} 
such that the embedded sequence produces greater classification accuracy.
The sequence of embedded terms $V_{1\leq i < M}$ is shared across all classes.
We adapt this to text and incorporate it into an end-to-end architecture.


\section{Data Preliminaries}
\label{sec:data}

For preference learning, we use the {\em Story Critique} dataset. The dataset consists of more than 80,000 unique stories with 1,378,696 total critiques. 
Every critique refers to a specific passage of the story, and so we construct 1,378,696 passage-critique pairs for training. 
The dataset is anonymized---unique identifiers including comment ID, submission IDs, URLs, and proper nouns have been removed. 
This {\em Story Critique} dataset was originally used to train the \carp{}~\cite{Matiana2021CutTC} contrastive model.

We also test our model on the {\em Moral Stories} dataset~\citep{emelin-etal-2021-moral}, to demonstrate our approach on datasets for which \carp{} was not originally trained.
The {\em Moral Stories} dataset consists of 12,000 short narratives. Each element of the dataset contains a context, a moral action, the consequences of that action and an immoral action and corresponding consequences.
Whereas the {\em Story Critique} dataset does not have critiques addressing the moral attributes of characters, we use Moral Stories to create a labeled dataset of stories and the {\em moral alignment} of the main character. Further where the moral stories dataset has only two label alignments we generate a dataset with three labels: ''good'', ''neutral'', and ''evil''.

%
Generation is done by randomly sampling a context, action and consequence and then few-shot prompting GPT-J-6B~\citep{gpt-j} to generate and then classify story segments as having a character that is acting ``good'', ``evil'', or ``neutral''.\footnote{The concept of character alignments is drawn from fantasy stories and role-playing games such as Dungeons \& Dragons~ \cite{si2021telling}. These labels are oversimplifications of the complexities that surround moral stance but allow for users to intuitively provide story critieria such as ``I want the character in my story to be good''.}
The logits associated with these labels provide a score for each label. 
In total, we produce a dataset with 17,157 story-alignment pairs.

To create a language model that generates stories that correspond with preferences
we first needed a language model that reliably generates stories/narratives.
We use the {\em ROCStories} ~\cite{mostafazadeh2016corpus} corpus, which consists of 100,000 five-sentence stories about common, everyday occurrences.
We fine-tune the GPT-2-750M language model on a subset of {\em ROCStories} to produce the base model from which all successive models are adapted.
Despite the model being trained on relatively simplistic stories, subsequent fine-tuning for preferences results in more expressive stories. This is due to the \carp{} model's tendency to shift the output distribution closer to human-written stories from the {\em Story Critique} corpus and {\em Moral Stories} corpus.
%
5\% of the {\em ROCStories} corpus is held out as a validation set, which is used later to assist with further fine-tuning.

\section{Fine-Tuning for Preferences with \carp{}}
\label{sec:ppo}



The base generation model---GPT-2-750M fine-tuned on {\em ROCStories}---cannot guarantee that any generated story will conform to a set of user preferences.
We look at the use case where a user may want to generate stories that meet some preference criteria such as being a sad story, having a lot of descriptive imagery, or involving a main character that has a good alignment.
Larger models---GPT-J-6B, GPT-NeoX-20B, or GPT-3---can accept more complicated prompts that include preferences in addition to the first sentence of the story.
However, these models are not required to attend to all elements in the prompt and, as will be demonstrated, can vary significantly in how much they adhere to the prompt.

Specialized story corpora for different preferences are not common. 
To tune a story generation model to meet given preference criteria, we require some means of judging generated story segments, computing loss, and back-propagating that loss back through the language model.
The \carp{} model can score a story segment based on how well a natural text target criteria applies. 
Casting a language model as a policy for generating the next event in an unfolding story,
we use Proximal Policy Optimization (PPO)~\cite{Schulman2017ProximalPO}) to fine-tune the base language model using \carp{} as a reward function.
PPO uses an experience replay buffer of tuples $\langle s_t, a_t, s_{t+1}, r\rangle$ where $s_t$ is a state at time $t$, $a_t$ is the action taken at time $t$, $s_{t+1}$ is the successor state, and $r$ is the reward earned. 
PPO samples batches from the experience replay and backpropagates loss through the generative model.
For tuning the language model we can think of the action $a_t$ as the next token to be generated given the state $s_t$ which is the sequence of previously generated tokens. The reward $r$ is given by \carp{} at the end of the trajectory (story) with the distance in log-probability from the base momentum model as an additional per token regualarizing reward.

To generate samples for the experience replay buffer, we randomly select a story from the held-out {\em ROCStories} validation set and use the first five tokens.
For preference learning, the best practice is to prompt the language model with a sequence within the task distribution instead of the \texttt{[SOS]} (start-of-sequence) token.
The successor state is the continuation generated by the story generation model.
The continuation is truncated at 60 tokens or the end-of-text token.
Finally, the reward component is the score generated by \carp{}, given the continuation and the target criteria text.

On every step, we sample 64 records from the experience replay.
The model is tuned for 20k steps, which takes on the range of an hour on a single A100 GPU. 
In preference learning, it is common to freeze all but a small number of the transformer blocks~\citep{frazier2019norms}. We found freezing all but the last two layers of the language model provided the best result.
Exact hyperparameters are provided in the appendix.





\section{Robust \carp{}}

\carp{} produces scores along a continuous range with many middling values. As a result\carp{} provides a relatively weak classification signal as observed during preliminary experiments. This makes it challenging for PPO to discriminate between continuations.
Consequently, the \carp{}-tuned model often fails to learn to meet some criteria and overfits for others depending on how sensitive \carp{} is to the target criterion.

Discretizing the reward scale and pushing contrastive reward scores farther apart has been a successful strategy in other text fine-tuning tasks (cf. \cite{Lu2022QuarkCT}).To make \carp{} more robust to generated continuations we adopt a similar approach, whereby we use a clustering-based pseudo-labeling technique to identify a discrete set of preference categories. Pseudo-labeling is a common technique used for image classification using ResNet~\citep{he2016deep} embeddings (cf. \citep{van2020scan,pham2021meta}). For a given story in the CritiqueCirclce dataset, a distribution over  the set of identified discrete preferences is computed as the softmaxed cosine similarity to each of the labels. This is used to form a new training set. In Section~\ref{sec:pseudo-labels} we overview the technique for identifying pseudo labels for criteria classes. In Section~\ref{sec:coop} we describe how to update \carp{} to use generated pseudo- and alignment labels.

\subsection{Pseudo Labeling}
\label{sec:pseudo-labels}

We generate pseudo-labels from the {\em Story Critique} dataset as follows. 
We observe that CARP's critique embeddings lie on a spherical manifold.
Thus we apply UMAP~\citep{mcinnes2018umap} to project the embeddings from their high dimension to 2 dimensions, which was determined via a hyperparameter sweep. 
The choice of 2 dimensions provides the added benefit of visualization of the \carp{} model's latent space.
Then we apply hierarchical density-based clustering (HDBSCAN)~\citep{mcinnes2017hdbscan} on CARP's critique embeddings to identify clusters of critiques.
Since HDBSCAN works best on vectors of low dimensionality projection to a low dimension is an important preprocessing step.

HDBSCAN resulted in 91 clusters. 
However, HDBSCAN failed to cluster half of the reviews it was given, which are subsequently added to a ``noise'' cluster.
%
We hand-label clusters by sampling several critiques from each. 
Any cluster where an associated story feature was ambiguous was discarded. Clusters with identical story features were merged.
As an example: there were two separate but nearby clusters associated with humor.
We observe that most of these clusters correspond 
with reviews of distinct story features---varying from the use of imagery, character dialogue, or humor. The full list of pseudo labels is in the appendix.

Finally we compute a high-dimensional centroid for every cluster by taking a sample-wise mean of the respective latent vectors. For a new, arbitrary latent vector, we measure its distribution over the classifiers as a softmax over its cosine similarities to each cluster centroid. However, taking the distance to centroid as a measure for classifying new points is often inconsistent for HDBSCAN.
For example, a maximal centroid similarity approach would misclassify points in the central cluster that are far from its center.
We alleviated this issue by removing samples with distance to centroid falling below a threshold.

\begin{figure}[t]
    \centering
    \includegraphics[width=0.49\linewidth]{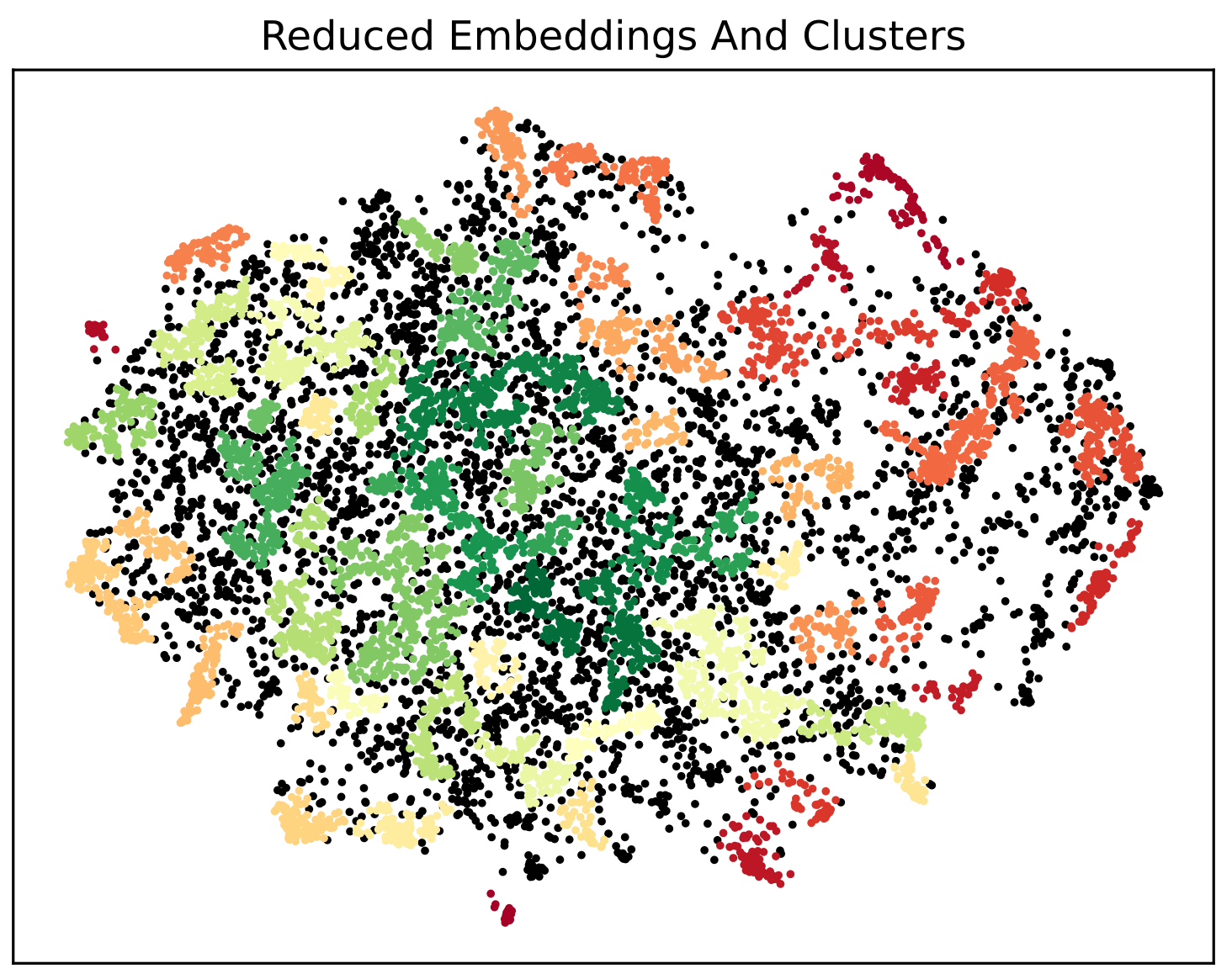}
    \includegraphics[width=0.49\linewidth]{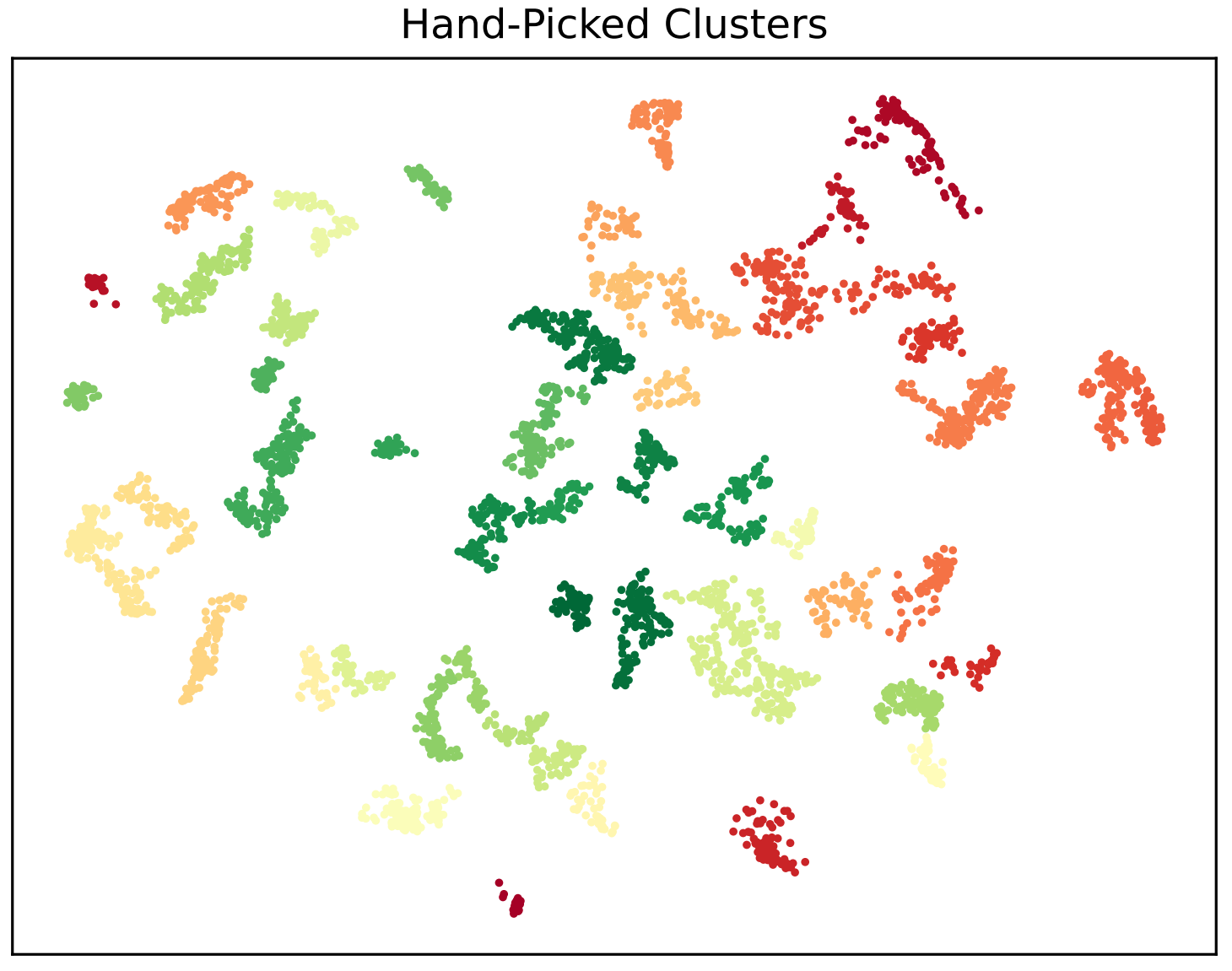}
    \caption{{\bf Left:} The latent space for CARP, represented with 2 dimensional reductions of the critique embeddings. Each point represents a review. Black correspond to points labelled as noise by clustering, while other colors correspond to cluster labels. 
    {\bf Right}: The hand-picked clusters.}
    \label{fig:carplatent}
\end{figure}


\subsection{\carp{} \coop{}}
\label{sec:coop}

\begin{figure*}[t]
    \centering
    \includegraphics[scale=0.3]{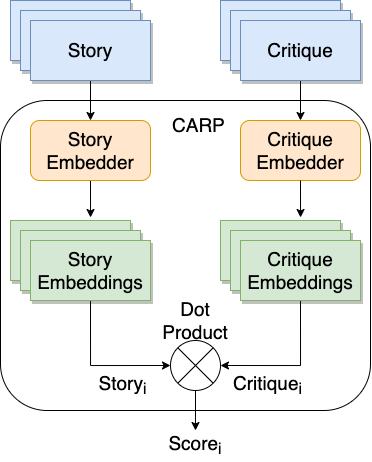}~~~~~~~~~~~~~~~~~~~~~
    \includegraphics[scale=0.3]{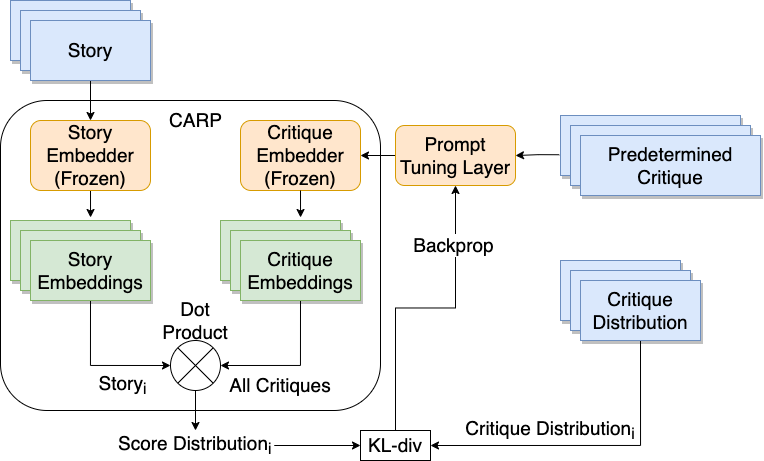}
    \caption{The \carp{} architecture (left) and the \carp{} \coop{} architecture (right).}
    \label{fig:carp}
\end{figure*}

In this section we describe how we update CARP to use pseudo labels.
Whereas the original CARP takes a story/critique pair and produces a cosine similarity score, we now require CARP to produce a score for each pseudo label, corresponding to each critique class. 

However, once we start using user-provided preference classes, we no longer have the text associated with the preference as an input. 
To rectify this issue, we simultaneously learn to generate a soft-critique for each critique class.
We incorporate the \coop{} prompt tuning technique into CARP in an end-to-end fashion as shown in Figure~\ref{fig:carp}(right).
Specifically, the \coop{} prompt tuning layer learns a \textit{unified embedding} \texttt{[V$_1$] [V$_2$] ... [V$_{M/2}$] [CLASS] [V$_{(M/2)+1}$] ... [V$_M$]} where \texttt{V$_1$,...,V$_M$} are shared parameters learned across all classes and \texttt{[CLASS]} is a token embedding that maximizes the log-likelihood of a specific critique class.

We train two new versions of \carp{}:\footnote{\carp{} \coop{} models tuned on pseudo labels and alignment labels are available at~\url{url redacted}}
\begin{itemize}
    \item {\bf \pseudo{}} is trained using pseudo labels derived from the {\em Story Critique} dataset as described in Section~\ref{sec:pseudo-labels}.
    \pseudo{} has six pseudo-labels derived from the Story Critiques dataset, chosen for semantic dissimilarity and separation in the embedding space. This choice allows for a stronger reward score during preference learning.
    \item {\bf \alignment{}} is trained using the augmented Moral Stories corpus described in Section~\ref{sec:data}.
    \alignment{} has the three alignment-critique embeddings.
\end{itemize}
Both models start with the pre-trained original \carp{} with embedding layers frozen.

%

To train \pseudo{} we filter the {\em Story Critique} stories on relative distance to the centroid of the desired pseudo label clusters, rejecting all samples with cosine similarity below twice the average cosine-similarity of the dataset. 
For the story/critique pairs that pass the filter we compute the distance of the critique to each centroid in the chosen clusters. 
We say a pair belongs to a cluster if it has minimal distance to that cluster among all clusters.
For the chosen critique pseudo labels, 
we select 1000 samples belonging to each class, balancing the dataset among pseudo-labels.
This is crucial as otherwise we observe the model overfits to overrepresented classes and fails to provide strong signals for less-represented classes.
Softmaxing gives a distribution over which we can minimize a KL-divergence loss between the predicted label and the target label. To train the \alignment{} model we perform a similar procedure as above, softmaxing over the logits generated by GPT-J-6B and thresholding. Here we observe it is sufficient to train both \carp{} \coop{} and \alignment{} on 1000 examples per label to achieve competitive downstream performance. This demonstrates the \coop{} method is highly data efficient, requiring a minimial number of examples per class to fine-tune when initialized with the pretrained original \carp{} model.



Figure~\ref{fig:carppipeline} shows our final process for fine-tuning the story generator language model with \carp{} \coop{} models. As before, the \carp{} \coop{} model produces scores for story segments generated by the language model. 
In this configuration, however, a criteria label is used instead of a text criteria and we use the negative log-likelihood loss for the corresponding label. Note the KL-divergence is computed as the distance to centroid as our stationary distribution (softmaxed distance to centroid per pseudo class). This is done to improve data efficiency since we found it was common for stories in the Story Critique dataset to fit under multiple critique labels.

\begin{table*}[t]
\scriptsize
\centering
\begin{tabular}{|c|p{0.45\linewidth}|p{0.45\linewidth}|}
\hline
& \multicolumn{1}{|c|}{\bf Imagery Sample Stories} &
\multicolumn{1}{|c|}{\bf Good Character Alignment Sample Stories}\\
\hline
\parbox[t]{2mm}{\multirow{1}{*}{\rotatebox[origin=c]{90}{\bf GeDi~~~~~~}}} & Her watch vibrated again, then stopped, her fingers moving over the glass. "Is it working?" she asked. "Yes," Harry said. "What do I have to do?" she asked. "Go back to the beginning," Harry said, looking at her through his glasses. After that, they could figure it out together. &
“Two, maybe three, years. We'll have to figure out what we want to do next." He pauses for a moment and then says: "We can make this work."
\\
\hline
\parbox[t]{2mm}{\multirow{1}{*}{\rotatebox[origin=c]{90}{\bf NeoX~~~~~~}}} & I was craving pickles, so I bought some for lunch today. I opened the jar, and I came to the pickles sitting in the middle of the table. In the middle of the pickles sat a huge bug. I watched it feasting on my pickles. It ate all of the ones on one side of the table and then ran back to the other side. I was so terrified and shocked by this. &
Yes, we  noticed that you accidentally deleted an important file two months ago. Oh, we also think that you're using ............... What's it called again? I believe it's one of those newly-invented compounds. It's been around for a couple of years, but I'm sure you've never heard of it. Of course, we're not threating you at all, but please take care. 
\\
\hline
\parbox[t]{2mm}{\multirow{3}{*}{\rotatebox[origin=c]{90}{\bf Default \carp{}~}}} & &\\
& It took him seven days. to sing three songs. on the first day he went to school in the parking lot someone opened the door. just sang the third time. the seventh time he sang. the one time someone jumped in front of him to stop him. &
Maybe Budge sensed my  sadness and anger. he walked over to me. he told me everything was alright. he said he was the reincarnation of a dead man. he was clearly just telling a story to make me feel better. \\
& &\\
\hline
\parbox[t]{2mm}{\multirow{3}{*}{\rotatebox[origin=c]{90}{\bf P/A \carp{} \coop{} }}} & &\\
& The creature eyed them curiously . Its sullen profile took on a glow, putrefying everything around it to a damnable criseness. It stretched its tentacles out into fishing web light, pinwheeling its obscene body over feet and head, its hooked fingers imperceptible in the darkness. Its tail and palpifer-like body protruded from fur. &
Hello, is everything  ok?  my son said one of his cats got loose and ran out the door.  we called the neighbor boy over, and he said he was going to look for it.   when he found it, the cat was stuck behind a fence.  that's when he came back saying the neighbor boy dropped the cat. \\
& & \\
\hline
\end{tabular}
\caption{Sample generated story segments.}
\label{tab:samples}
\end{table*}

\section{Experimental Design}

We test our pipeline in two separate regimes: First with access to labeled preference data with which we can fine-tune \coop{} and GeDi models. The second case is without access to labeled preference data. Here we leverage the Pseudo \coop{} model and the pretrained GeDi model to guide models generating text adhering to a particular topic. 

We recruit human study participants and ask them to choose which preference from a list best describes a segment of generated story. In the case we have labeled preference data we use the alignment dataset labeling main characters as being either \texttt{good}, \texttt{neutral}, or \texttt{evil}. We call these fineituned models Alignment \coop{} and Alignment GeDi, respectively. Otherwise participants select topic labels from among \texttt{family}, \texttt{music}, \texttt{accidents}, \texttt{religion}, \texttt{imagery} (involving a lot of descriptiveness of visual properties), or \texttt{fighting}. Often stories adhering to a certain alignment often involve a specific topic, e.g. good stories often center around family and bad stories often involve accidents. Hence we separate the evaluation of stories generated with a desired alignment from the those with a desired topic to prevent artificial mislabeling.

In total we evaluate four separate classes of models:


\begin{itemize}
    \item{\bf GeDi LM:} The GPT-2-750M language model guided via GeDi.
    
    \item {\bf NeoX:}
    The GPT-NeoX-20B language model prompted with the preference criteria and an initial sentence.
    
    \item {\bf \defaultlm{}:}
    The base GPT-2-750M language model fine-tuned via vanilla \carp{} rewards (as described in Section~\ref{sec:ppo}. 
    This model is an ablation of the full model to assess the importance of \carp{} \coop{}.

    \item {\bf \carp{} \coop{} LM:}
    The base GPT-2-750M model fine-tuned via \carp{} \coop{} rewards.
\end{itemize}

This results in 36 different story generators
(6 topic labels $\times$ 4 models, plus 3 alignment labels from augmented {\em Moral Stories} $\times$ 4 models). 

We recruited 25 people on the Prolific\footnote{\url{https://prolific.co}} crowdsourcing site. 
Each participant read 44 story segments drawn randomly from the set of generated story segments across all models.
Participants took on average 18 minutes to complete the task and were paid \$12.00 per hour. 
Stories are generated by randomly selecting 5 prefix tokens from a held out ROCStories validation set and generating the continuation. 
In total, we generated 20 story segments per preference per model.
All generated stories are available in the appendix.



\section{Results and Analysis}


\begin{table}[t]
\centering
\footnotesize
\begin{tabular}{|c||c|c|c|c|}
 \hline
{\bf Criteria} & {\bf GeDi} & {\bf NeoX} & {\bf \carp{}} & {\bf \carp{} \coop{}}\\
\hline
Topics & 0.371 & 0.489 & 0.533 & {\bf 0.615}\\
\hline
Align. & 0.561 & 0.460 & 0.506 & {\bf 0.675}\\
 \hline 
\end{tabular}
\caption{The average selected human preference across all criteria for each model.}
\label{tab:overall-pref}
\end{table}

Table~\ref{tab:overall-pref} shows the aggregate percentage of the time participants correctly selected the correct criteria for each story segment read, broken down according to whether the particpants were choosing among topic labels or character alignments.
It shows overall that participants were better at identifying the correct criteria when reading story segments generated by \carp{} \coop{} guided LMs.

\begin{figure}[t]
    \centering
    \includegraphics[width=1.1\linewidth]{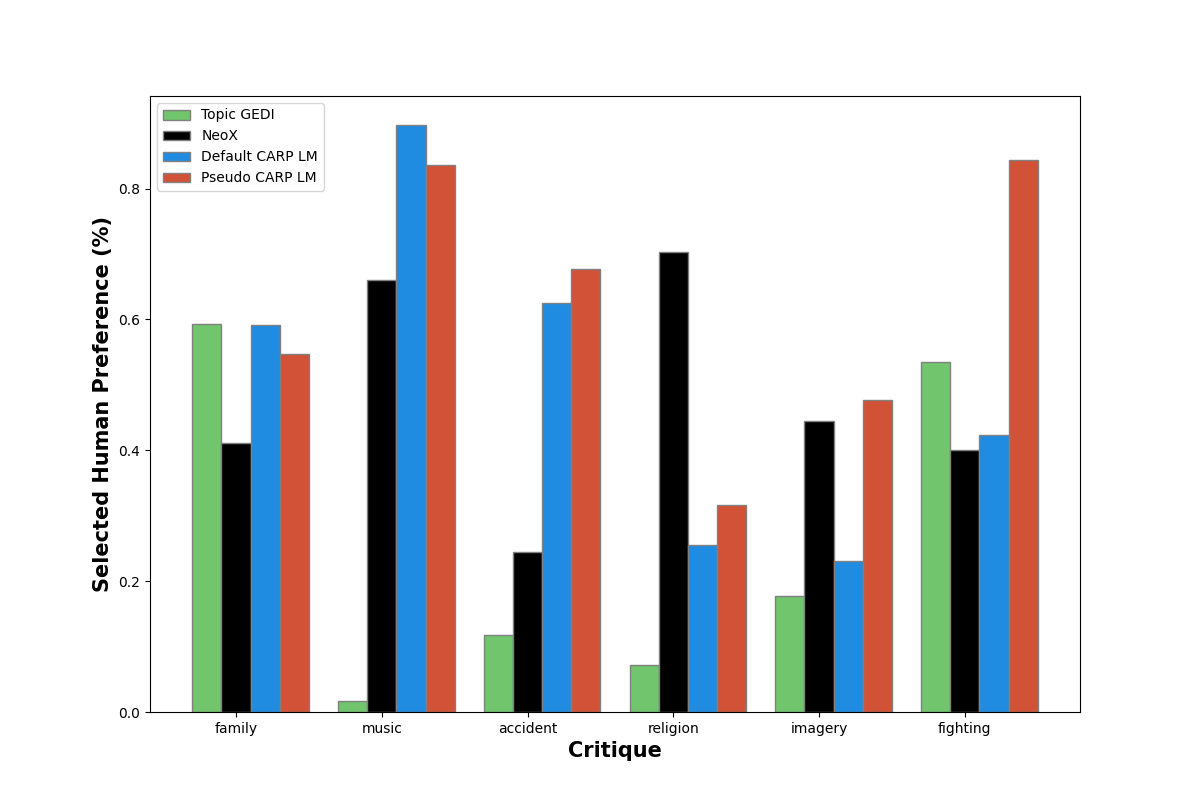}
    \caption{Histogram of human preference for GeDi, Neox baselines, Default CARP guided LM, and Pseudo CARP guided LM over critiques. Note we we do not have pseudo-labels for romance, horror so do not evaluate a Pseudo CARP guided LM on these critiques.}
    \label{fig:critiquestats}
\end{figure}

Figure~\ref{fig:critiquestats} shows how often people identified the preference criterion used to generate each story across models.
The degree to which participants could identify the criterion used to prompt NeoX varied widely. 
In some cases, CARP guided LMs performed better than NeoX, but not universally so.
\pseudolm{} beat NeoX in creating stories with recognizable criteria in all labels except ``religious'' stories for which it seems to be particularly sensitive. Pretrained-GeDi guided models do decently well on some topics despite not being included in the original GeDi training set. However in other topics GeDi does quite poorly, particularly with stories involving action. In all cases except \texttt{religion} \carp{} \coop{} guided models are preferred.

\begin{figure}[t]
    \centering
    \includegraphics[width=\linewidth]{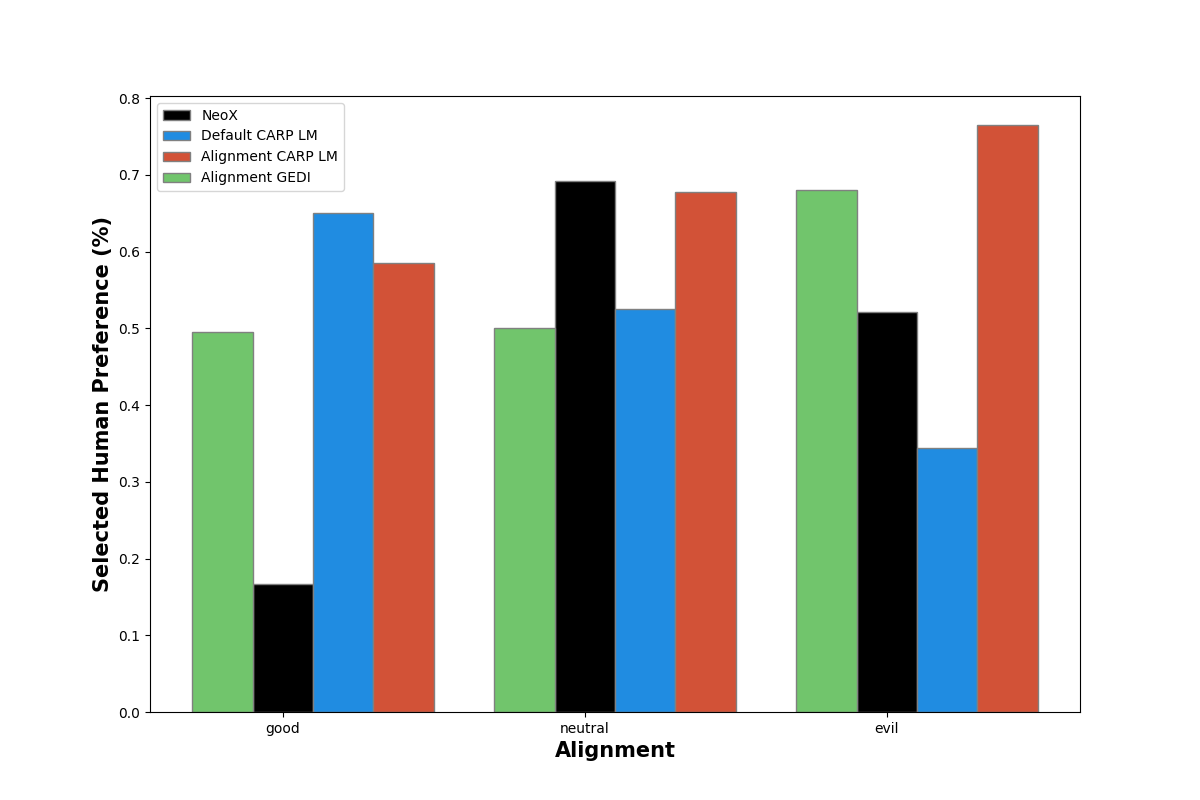}
    \caption{Histogram of human preference for GeDi, Neox baseline, Default CARP guided LM, and Alignment CARP guided LM over alignments.}
    \label{fig:alignmentstats}
\end{figure}

Figure~\ref{fig:alignmentstats} shows results on the character alignments criteria.
As before, NeoX has widely varying quality of response to the different criteria. For all three criteria, \alignmentlm{} meets or exceeds NeoX. CARP guided LMs only exceed NeoX on the ``good'' criterion. GeDi guided LMs perform decently well, despite the complexity of the task, when GeDi fine-tuned with the Alignment dataset. However on average \carp{} \coop{} is preferred.



The NeoX baseline is many times larger in terms of parameters than the other models fine-tuned with different versions of CARP below. The NeoX baseline also has direct access to the criterion as part of the prompt, whereas other models have the criterion implicitly represented. Yet despite these advantages, NeoX fails to preffered to \carp{} and \carp{} \coop{} guided methods a majority of the time. We conclude that NeoX, despite its size, is not guaranteed to attend to prompts with preference criteria in all cases.
A generative model over 20x as small and tuned \carp{} rewards proves to be more reliable in its ability to provide stories that meet given criteria. These models are made even more robust by discretizing the criteria into distinct classes via \carp{} \coop{} and learning a prompt that generates a stronger, less ambiguous, reward signal.

Additionally of interest is the observation \pseudolm{} and \alignmentlm{} generative models both start from a base model tuned on ROCStories, which are simple stories with simple sentence structures. Without ever seeing an example story from the {\em Story Critiques}  or {\em Moral Stories} datasets, these models learn through reinforcement by \pseudo{} and \alignment{}, respectively, to prefer language and story structures that are more complex than the original ROCStories dataset. 

\begin{figure}[t]
    \centering
    \includegraphics[width=\linewidth]{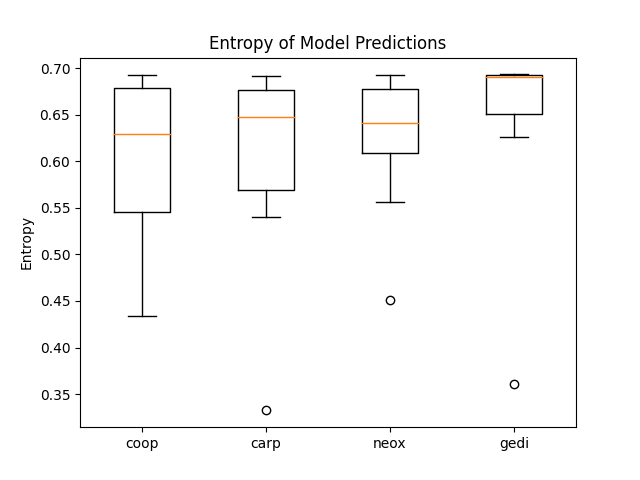}
    \caption{Entropy box-whisker plot of model generations annotations. Lower is better. Entropy is computed as the binary entropy over human participant answer proportions.}
    \label{fig:entropy}
\end{figure}

Finally we observe an average inter-annotator agreement on questions of $0.74$ which is decent despite the complexity and potential for multi-label applicability in the tasks. Agreement is around $0.77 \pm .01$ among model classes except GeDi guided LMs which have noticeably lower agreement at $0.62$. We speculate this is due to GeDi's poor performance for some out of distribtion topics. Similarly agreement on questions within each preference class is $0.75 \pm 0.04$ with the exception of the \texttt{imagery} preference which is low at $0.58$. We speculate this is due to the complexity of the task. These observations are supported by an entropy test (figure \ref{fig:entropy}) which measures annotator agreement across models.

\section{Limitations}

The dataset used to train \carp{} primarily uses short-form creative writing samples and in-line critiques of these samples. One primary limitation of the model is the inability to reason over longer sequences of text. As such, the model also cannot reason effectively about preferences or critiques as would be applicable to longer form creative writing.
Like other neural text generation tasks, longer stories would be generated by successively generating segments and pre-pending to the context prompt.
In such a case it would be feasible to change the preference or criteria mid-stream.

Since \carp{} is trained solely on short form stories and in-line critiques of those stories, we lack the ability to reason about sequential or long form preferences. In a similar vein, it is not possible to condition the model based on sequential forms of human preferences (e.g. "This story would be better if it were X, and so in turn I would change Y or Z in the writing or plot structure"). Tuning to the preferences of an individual editor or reader is also not possible.

The experiments focus on six of the more well-bounded and separated critique classes in \carp{} embedding space. 
As such, our results represent an ``upper-bound'' of sorts regarding downstream language model performance.
However, all 91 clusters are distinct enough that were we to train \pseudo{} on all pseudo labels, the embeddings would not change much from the original \carp{}.
That is, \pseudo{} generally improves performance or performance is unchanged but rarely ever hurt by additional pseudo-label training.

The Story-Critique dataset is non-public due to licensing issues despite best efforts to make it publicly available. As such, exploring the preferential biases which are undoubtedly present in \carp{} is not feasible for these authors though we do acknowledge the complication this presents in evaluation, assessment of ethical concerns and general limitations of the model/approach.

\section{Conclusions}

We have demonstrated a GPT-2-750M can be fine-tuned using reinforcement learning via 
contrastive model trained to correlate story segments with human text critiques.
With a novel combination of pseudo labeling and integrated prompt tuning techniques, this fine-tuning technique becomes more robust because the reward signal becomes stronger and more discriminative.
The resulting models are, in nearly all cases, competitive with or outperform  20x larger prompted models when it comes to observed conformance to preference and character alignment guidance.

Our results are not unique to \carp{} and the {\em Story Critiques} corpus on which it was trained. 
We show how the {\em Moral Stories} dataset can be augmented to provide story control guidance revolving around the moral alignment of characters. 
The models trained from this are significantly more capable than larger models that rely strictly on a model's ability to attend to instructions in the prompt.
It points to future directions for aligning models with human value systems.

The work presented provides a means by which to give users of generative neural language models control over certain qualities of the outputs through the use of general purpose preference models. This makes language models more useful for human-initiated creative activities such as fictional story writing, but also provides a means of control for other generative text process.


\section*{Ethics Statement}

As with all neural text generation systems, our work is prone to echoing the biases present in the dataset \cite{sheng2019woman} and generate non-normative text (i.e. in violation of social norms).
No existing automated storytelling systems is able to entirely eliminate these biases.
Our system is susceptible to biases in the {\em Story Critique} dataset.
In preliminary experiments the model often conflated extreme acts of violence with humor.
However, in some cases, the ability to provide guiding preferences can reduce the amount of toxic, prejudicially biased, and non-normative output because the preferences can push the model into latent spaces where these are less prevalent.

With large scale language models, out of distribution (OOD) inputs or preferences present a challenge. If exemplar stories are not present in the training set, they will not be well understood by the model nor will a prompt or critique mentioning these topics perform well. For example, if most training data concern cisgendered, male vampires, then the model will struggle to provide representative stories and critiques about queer, non-binary vampires.

Fictional stories that are presented to readers as non-fictional can be used to influence or misinform.
We, and others who work on story generation, must take care to ensure that generated content is not presented as factual real-world occurrences.  

The ability to control neural text generation output---whether for stories or otherwise---may have important downstream applicability.
In particular, the ability to control generation makes text generators less unpredictable for writing assistance.

\bibliography{anthology, main}
\bibliographystyle{acl_natbib}

\newpage
\appendix

\section{Preference Learning Hyperparameters}

Below we've included an example of CARP CoOp preference learning parameters.

\begin{lstlisting}
{
  #LM Model args
  "lm_name": "gpt2-large",
  "ref_lm_name": "gpt2-large",
  "tk_name": "gpt2-large",
  "num_layers_unfrozen": 2,
  "save_folder": 'ckpts/roc_evil_coop_model/',
  "use_lm_ckpt": True,
  "lm_ckpt_path": "raw-roc-gpt2-large/",

  #Carp model args
  "carp_version": "coop",
  "carp_config_path":...
  "carp_ckpt_path": ...

  #Training args
  "steps": 20000,

  #PPO Args
  "ppo_epochs": 4,
  "txt_in_len": 14,
  "txt_out_len": 60,
  "lr": .5e-6,
  "init_kl_coef":0.2,
  "target": 50,   #KL Divergence target
  "horizon":10000,
  "gamma":1,   #Discount factor
  "lam":0.95,
  "cliprange": .2,
  "cliprange_value":.2,
  "vf_coef":.15,

  #Review
  "review": "evil",

  #Dataset
  "data_path": "dataset/roc_prompts.txt",

  #Minimize or maximize
  'minimize': False,
}
\end{lstlisting}

\section{Story Examples}

\begin{table*}[t]
\fontsize{8}{8}\selectfont
\begin{tabular}{|p{0.3\linewidth}|p{0.3\linewidth}|p{0.3\linewidth}|}
\hline
\multicolumn{1}{|c|}{\bf NeoX Stories - Good} &
\multicolumn{1}{|c|}{\bf Default \carp{} Stories - Good} & \multicolumn{1}{|c|}{\bf Alignment \carp{} Stories - Good}\\
\hline
What do I do? I wonder. Each fly buzzes around trap to trap, ignorant of all the other traps. How many do I have to be? He looked at the back of the bottle trying to get a view of the drafty air duct. That's right, you knew it. You can hear me. I'm on to you. I knew you had a compliment hidden in there. Why else would you listen to  &
she saw a homeless man in the park outside her home. she felt bad enough to buy food and shelter for him.  but when he asked for money, she realized he needed more help.  she took him to a shelter last night and worked to help him in return. & 
What do I do? i wonder how else can i make myself happy?  after all, life is a constant journey.   today, i invite my cousin to join me.  we have a great time exploring the countryside together. tomorrow we both hope we can join in another adventure!
\\
\hline
What have you done today? When you were little, you baked your first cake for your birthday. The results were a big success. Gathered around the table...most of the family will take a piece. Everyone...was pleasantly surprised at how good they tasted. And the best part? Before the icing was even done? You licked the bowl. But the tears came as swift as the voice in your head:  &
Yeah, don and i went to the same high school.  we played on the same soccer team, and i won the first game.  the second game was a tough one.  in the end, i'm not as good as don, and i lost.   we won the championship, though. & 
“What have you  done today?  everyone, including me, is happy for today.   we worked overtime, but it was worth it.  we finished all of our assignments.  we had a lot to do, but we all did a present first.
\\
\hline
 You're missing a report your boss told you to file. You have three chances to turn it in, or you'll lose your job. He is waiting for you in his office. He'll have to take no for an answer. You take the chance and hand it over. "Thx, boss!" You hit the eject button on the CD player. You've got 99 problems, and your stereo's not one &
Okay, what's the job  for a homeless man who begs everyday?  most people say watch his own back, and some donate money.  but looking outside his window last night worried him. a homeless man sat among the tall trees looking quite miserable.  before he could cross the street, a car made a scare worthy of tale to tell. & 
You're missing a head band for science camp. we have an idea. we should make one out of string. string from fishing distractions. that way students can dance all month.
\\
\hline
The old man stared at the worn, ragged, tattered clothes hanging in the closet. He sighed. "Where did I get these?" He asked himself for the thousandth time. He remembered that he had gone to the lost and found. He took them out. All of the other clothes on the rack were in decent shape. They were all new. They were all appropriate. Wouldn't it be nice if  &
It's not a stupid ball  to pick for a girl. you know it makes her likes you. so you pick one. and then you look for the next. until she's on you. & 
The old man stared breath lessly at his granddaughter. his granddaughter looked up from the floor.  she held out a hand, and he gently grasped it.   he smiled, and said he liked her a lot. 
\\
\hline
John0 lifted his Springfield up to his ear. The executive office suite was in fine working order. "Hello, again, Mr. Squawk," John whispered to his new toy, which beeped in response. If only John knew...SPLAT! "Crap," he shouted. He reached into his desk to grab his earplugs and clean them off, but only managed to tear off his must 
 &
This is a story about a man who saved a stranger's life. he said he didn't need money or a car & 
I hugged him closer. It was good that i hugged him back. this felt good. i could feel my happiness growing. we went inside and enjoyed each other's company.
\\
\hline
Please hand me the hat on the table. - >Good /?-The-pig-pees-on-the-carpenter's-shoe-and-the-hammer-is-lost.-!-Excellent-!-I-have-one-question-:-(A\%morality-question)What-does-good-behavior-get-the-pig-in  &
He took a breath. he saw the homeless man sitting on the park bench. he walked to him and gave him money and gum. the homeless man thanked him and walked away. he saw another homeless man being harassed and followed this one. & 
Jane0 heard a sudden  howling noise from the attic. she awoke to find her cat pawing at a box of books on the floor.  the cat was scared and flinched, but soon calmed down.   it seemed as though months had passed, and then years.  then the cat whined and went back to sleep.

\\
\hline
Get Jane0 into her aikido class; she's already five minutes late. She likes the aikido class, and she's been working very hard on katas. The class starts only an hour later than normal, and she's here early, under the venetian blinds in the waiting area. The door opens, Jane0 steps in, and belatedly fills in a registration form. Inside &
You asked for it .  when it was mine, i gave it to you.   when i told my parents, my parents denied owning it.  that dog died. only i was able to take it to a homeless shelter. & 
I admired Melaney for being so committed to giving away free food.  in fact, melaney encouraged me to join her in her mission.  i decided to join her organization and give away free food to needy kids.  every day, i painstakingly package and distribute free food to others.   at the end of the year, i donated over 700.

\\
\hline
What do I do? I have a problem that I have been worrying about. I am really upset by the problem, and I cannot keep from thinking about it. It is like a voice in my head that is always talking to me. The voice tells me that I should do something about the problem.Now what am I supposed to do? As much as I'd like to think that the voice is right, I 
 &
Maybe he wasn’t  was a homeless man.  he never seemed to be able to find employment, including his own.   one day, a man noticed his coat was missing.  the can was believed to be responsible for his condition.  and sadly, he died before he could be found. & 
I know,  right?  we were on the playground. my friends and i decided to jump on the slide. the happened to fall straight to the bottom. we all laughed and went back on the slide.

\\
\hline
the girl's friend was upset when the boy apologized. She didn't like him apologizing and making the only one in the car (the girl) feel bad. What the girl's friend didn't mind was his explaining to the PILOTS how he felt and then the pilot asking the girl to turn on the radio so that they could hit the 10,000-feet mark and let them know that they were okay.  &
John0 get off him  couch and walked to the kitchen to make dinner. he removed the cat food from the refrigerator. he found that water had overflowed the container. john0 decided to drain the water out of the container. he now has cat food and water in the pantry. & 
I shall study the dolls called since this term has begun.  the first time i saw them, i thought they were real.   seeing their effects on children that year, i changed my mind.   i saw them again this year, and although their effects were real, i chose compassion.   nowadays, they are just props

\\
\hline
Is your dad a magician or something? He moved the box from off my bed, to the edge, and then back again.He's quick! He's quick! I want to be able to do that. .......................... I think I'll try to count... let's see... one elephant, two elephant, three elephant... nope. Still eight.O-kay. What was the question again?"Dad,  &
I found the perfect angle  for a portrait. the man was very respectful of my work. he smiled and helped me with the work. he took a picture with his own private camera. he gave it to his friends for memories. & 
Was this the puppet master doing all his puppet shows? he would get up early in the morning making sure his puppets were perfect. then set them up in clubs across the neighborhood. right before school everyone would line up for the show.

\\
\hline
\end{tabular}
\label{tab:samples}
\end{table*}

\newpage 

\begin{table*}[t]
\fontsize{8}{8}\selectfont
\begin{tabular}{|p{0.3\linewidth}|p{0.3\linewidth}|p{0.3\linewidth}|}
\hline
\multicolumn{1}{|c|}{\bf NeoX Stories - neutral} &
\multicolumn{1}{|c|}{\bf Default \carp{} Stories -neutral} & \multicolumn{1}{|c|}{\bf Alignment \carp{} Stories - neutral}\\
 \hline
Abhar- dur was performing a complicated calculation on the blackboard when Sigurjon gave him a nudge. "If I were you, I'd stop and leave a space." "It's time for a break, anyway." replied the teacher, breathing hard. He was a thin man with unusually large ears, and his frizzy hair was thinning a bit, too. Abhar-dur finished the   &
John0 unclipped  the parachute cord. the plane was going to fly in a different direction. test pilot paul controlled the plane. they were in the air over texas. it was a flight of tragedy. &
Abhar- dur was performing a kazoo. betty hid behind him. she got out of his seat. she ran towards the floor. i hope she doesn't embarrass him.
\\
\hline
He shook his head. "This is ridiculous, Sis. If I'm hosting, I'm leaving.Come on, we'll catch up with them later."Can you identify which statement in this passage is the protagonist's ruled attitude, and which is the protagonist's opposites? Choose the best answer. [A] B [B] C [C] D [D] E [E]   &
He smiled briefly at her . then he reiterated something he said earlier. he laid out several scenarios. they ranged from minor inconveniences to very serious ones.  in the end, he told her to relax. &
He shook his head. he muttered under his breath as he walked down the block. he decided today was not a good day to go outside. he couldn't believe he even got that far today. walking home from work he could feel rain drenching him.
\\
\hline
The bear noises on this example have no meaning to this character. The character might not have any unique sensory experience with the river, for instance. Such behavior might be indicative of a person not taking their daily routine seriously. Or, the character might have seen a bear in the woods and be so frightened that the sounds of a river are being imagined. This, in a way, is a trait, not a neutral   &
Mr Whoul an <UNK> was a character in disney world. he was a line bearer for multiverses. a person would ask a strong question and the character would respond. the audience gets a good chuckle when this happens. they <UNK> how funny it is that this show has characters like this. &
The bear noises are familiar  to janet. every year the same noises start from the trees. janet sits outside in an old chair looking out over the trees. she leaves the chair to go inside when the noises stop. janet doesn't come back to sit in that chair anymore.
\\
\hline
I shook my head. "I am curious, but that's not why I kept hitting myself."The answer seemed obvious. "I was trying to burn the poker," I said. There was a pause."Why?" a voice asked. I knew that voice.Axl Rose."Because I wanted to kill you!" I yelled. I swung the poker into his knee. The leg crumpled. He tipped over   &
The man shook his head . the woman shook her head too. the man asked why. the woman shook her head too. the man wondered why they were doing that. &
I shook my head. someone called from the break room. someone said the soda came from the vending machine. someone said they put the soda in there by mistake. someone gave me a minute to drink it
\\
\hline
John0 waved his hand in the air. "Maybe so," he said laughing, "but I like to think that I smile when I make a deal--that I have a perpetual itch to make a deal. This is my style. Deal making is in my blood." "However, I will smile when I reach my goal."  &
Yvl's  admirers saw fit to pay him some lip service. they simply said they knew what a man was.  good for them, because many would-be Pillar veterans still feared him.   good for them, because having his life publicly dissected was gratifying.  it was also gratifying that he had been chosen as the &
Of course,  i had to buy some new pants.  they were grilling season and the forecast looked grim.  i looked on the internet, but i didn't have anything cooking.   instead, i went to the local farmer's market.   there, i found nice greens and good-quality goat cheese.
\\
\hline
Do that scorching-hot poker make you think of your granny or your scoliosis?Chances are, you've never thought about either one of them.According to a new study, we choose a relatively small number of what are known as prototypical responses to tell our brains what to expect in a given situation.Researchers also believe that the capacity   &
What would they do next ?  the president told them to become intelligent, inventive, and hard-working.   they each bought a blackboard, a rubber stamp, and a notebook.  they wrote down their ideas daily.  after a month, they had written down their mostiled ideas. &
Don'  was watching a movie with his girlfriend. they needed to rent a movie. don eventually suggested taking advantage of a software feature. his girlfriend liked the idea. they used the software feature to pick a movie.
\\
\hline
Nekolai nodded and the two shook hands. Bellatrix saw she was looking young. She couldn't be more than six. "He is my friend," she said. "But he is very sick just now." She took his clammy hand in her own and squeezed. "But he is getting better." She let go. "You can come see him. Come on, Nekolai" (2   &
The men are shocked by their political affiliations. many of them are actively campaigning.  when they all vote, they often have strongly held beliefs.  some voters represent their strongly held beliefs the best.  from their experience, they learn not to stray from their instinct. &
Oh, come on,  do you hear me?   you know the drill, don project.  you know, that titanic fail dish that you mother made you wear?  some watch channels have it on the newest reboot.  so as soon as i found this dish, i refolded my mother's outfit.
\\
\hline
He picked his words carefully. 'We had better wait, and find out if it's a mistake. If it's not, I think it's a great idea.' He didn't feel enthusiastic. It was a big risk, after all. They had a lot riding on it. And if it didn't work, if it fell through... They'd be stuck. No money. No money, of course. But they   &
Still too scared to answer  any questions jones had, the interrogator did nothing.  at one moment jones realized he might be in trouble. the interrogation had not been <UNK> as there was no conclusive proof. in a fit of rage jones demanded to know for sure what was on the tape.  nothing but rage, jones knew he would have &
I'm  determined to fix this mess. i go to the garage to grab my tools and start work.  after putting the parts together, i start work on the brakes.  it's now my job to make the vibrator stop squeaking. i'm tired but satisfied that i finally fixed this car.
\\
\hline
Jane0 looked down at her hands and sighed. She was bored. The book she was reading was so predictable; all Jane could do would be to reread the same paragraphs over and over and over. She had read it six times now, and had heard the same passage repeated at least four times. "It's not willpower," she thought as she took another sip of lemonade. "It's not willpower   &
At these words, Sam  met the love of his life.  they chatted about life for three days. by day five they were passionately in love.  though he was nervous, he proposed to her that night.  they married that same night. &
There's  a new chocolate store in town. we searched but couldn't find it. finally someone asked if they could post a photo of the store.  relieved, we gave them the photo.   finally, the store had been photographed.
\\
\hline
I snap out of it, and the girl I stare at continues to dig out her earbud. As I try to channel my brother's advice, I slowly approach her. She gets even more distant, and all of a sudden she removes her hood and I realize she's not wearing a shirt. I back away, and she hollers something at me. I don't understand what she's saying. She steps towards  &
"You' . you talk like a bull. i wish you would stop talking. i hear you calling me a baby. i wish you would stop talking. &
I said sit down . he said no. i wanted to chew something. he finally told me to stop chewing. i had no choice but to sit back down.
\\
\hline
\end{tabular}
\label{tab:samples}
\end{table*}

\newpage

\begin{table*}[t]
\fontsize{8}{8}\selectfont
\begin{tabular}{|p{0.3\linewidth}|p{0.3\linewidth}|p{0.3\linewidth}|}
\hline
\multicolumn{1}{|c|}{\bf NeoX Stories - evil} &
\multicolumn{1}{|c|}{\bf Default \carp{} Stories -evil} & \multicolumn{1}{|c|}{\bf Alignment \carp{} Stories - evil}\\
 \hline
John0 grabbed the doll from the toddler's hands and threw it across the room. The doll fell on the bear, which was named Leopold. The blood drained from the bear's face and it lay on the rug in a heavy, inert mass. His right paw was still clutched in the thing's death grip. John couldn't look at it. "The hell...?" One hand covered his eyes. He   &
Sam0's first impression  of a woman was her emotionless face. so when he met the woman he was stunned. she was cold and aloof. sam was stunned by her behavior. he asked her why she acted that way. &
John0 grabbed his elbow . another student confronted him about it. the student beat up john very badly. john punched the student and punched the teacher also. john was suspended as a result.
\\
\hline
John0 help me, I'm starving!I haven't eaten for days! My name is John. I just turned 25. I have less than \$500 in the bank, which I need to pay for my rent, electric, telephone and food. I'm a janitor and my neighbors threw a party for my birthday. I drank a "long island iced te" and had two "whites"   &
The room was empty. the beggar sat on the street. people seemed very sad for him. in fact, they cheered when they gave him money.  the beggar laughed and said that they only felt sorry.  he gave back more money than he ever earned, until he died. &
John0 help me,"  he shouted to no one in particular.  his presence caused everyone else to take notice and think.  when a small girl came to john0, he threatened her.   instead of kicking her away, john0 tackled her and pinned her to the ground.
\\
\hline
Who' s there? Is something the matter? yea, right? I’m looking for a clothesline. You know, where you can hang your clothes on. Limegreen shirt, red tie, navy blue pants. Blue striped shirt, brown pants, green tie. How did you know my closet needs a make-over?... I found one at your place. Thanks for the tip. Its   &
I know,  right, that evil must have lived among us.  evil seemed like it didn't want to exist. then one day evil dragged itself to a town. there it all was in all its glory.  the people turned out in force, and evil was banished forever. &
"Who' s there?  i asked my son, who's there?  he didn't reply.  i turned around, and walking toward the sound.   just then, my neighbor stepped out from behind a bush.
\\
\hline
Yes, we  noticed that you accidentally deleted an important file two months ago. Oh, we also think that you're using ............... What's it called again? I believe it's one of those newly-invented compounds. It's been around for a couple of years, but I'm sure you've never heard of it. Of course, we're not threating you at all, but please take care   &
This story was about an evil clown. i don't think people who do evil things will get any praise.  i really do think evil people are never rewarded. &
"Yes, we  noticed that the dog was eating her breakfast.   since she loves toast, we attempted to feed her toast.   after a few minutes, she instantly after spit it out.   baffled, we asked her what's wrong.  she stated that she hates toast and how can possible
\\
\hline
Lady Pember was the mistress of Lord Pem. (The narrator is specific as to where the novel took place, but the material here is entirely realistic.) Lady Pember at that time being in her mansion. Lord Pem, coming to give her the money, means to steal it. He obtains a gag against her jaws from one of the servants. He inveigles her to open her mouth with the   &
This story was about an evil man.  he was at the local library during a movie night.   evil person in the dark got out of an old it movie and gave them a scare. &
"Lady Pember  was jailed for armed robbery.  when her other suspected assailant didn't show, she framed him.   before he walked free, she shot him in the head with a 9mm gun.   he went into coma, but recovered after a few days.
\end{tabular}
\label{tab:samples}
\end{table*}

\newpage 

\begin{table*}[t]
\fontsize{8}{8}\selectfont
\begin{tabular}{|p{0.3\linewidth}|p{0.3\linewidth}|p{0.3\linewidth}|}
\hline
\multicolumn{1}{|c|}{\bf NeoX Stories - fighting} &
\multicolumn{1}{|c|}{\bf Default \carp{} Stories -fighting} & \multicolumn{1}{|c|}{\bf Pseudo \carp{} Stories - fighting}\\
 \hline
They began to exchange blows. Peter's blow was so hard that Danny was knocked out of the ring. After eight or ten blows were exchanged, Danny finally went down. Peter followed him down and began punching Danny in the head. He managed to put Danny's left eye out of the socket, and Danny was beginning to lose balance. Peter grabbed Danny by the head and neck and manhandled him  &
"No!", she screamed as she ran towards him.  he reached out and caught her head.  she woke up, screaming wildly.   she said her sorry, and he punched her in the gut again. &
They cleared the lower level  of the forest.  they swung the ax, missing the tree.   with one mighty swing, one branch fell with a tremendous crash.  a horse stood at the impact site with its head hanging over a hole.  the horse panicked, and charged at the bandit.
\\
\hline
I smiled as I looked upon him in the arena. All the blood had drained from his face and the limb that still held the spear was shaking. I think my smile must have been terrifying; Hephaestion was quite still. His shield had fallen to the ground and in his hand he held only the wooden shaft of his spear. Wide-eyed with horror, he glanced to the left and broadened, frozen  &
John0 reached underneath his  desk. a gun fell out from its holster.  before anyone can react, he took aim.  he emptied his entire magazines into his target.  after he hit his target, he walked away. &
I smiled as I stood  and swung his bat. he swung and swung and hit me in the shoulder.  before i could get up or get on my feet, he swung again.  i stumbled backwards and ended up on the ground game.  i looked up, caught my balance, and managed to escape.
\\
\hline
Neil had just finished refueling its single-engine cargo plane when the first patrol passed. Taking off from open water a few miles away, the two boats circled to approach from behind, guns and rockets blasting. Neil's work over Iraq, flying ground attack missions and flying covert insertions of commandos, had taught its pilots well. During one firefight over Northern Iraq, their plane had been shot down  &
Now you listen to me  roughly. i started to walk towards the car. someone grabbed my hood. i turned around and pushed them away. my friend was chasing behind me. &
John0 scowled  at his opponent.  he swung his bat, and knocked the opponent out.  johnyon defended his stance.  he launched a soft knee at his opponent, knocking him out.  the opponent fell on the floor.
\\
\hline
I drove home from work at 100 clicks an hour. 'Come, sit by my side, you can drive with me, you'll see,' he said. He is offering you a ride. I speeded up. He said, 'Don't be scared! Drive fast!' He said, 'I'll be there for you'. But, he probably wasn't. They always give you the  &
Tyres screeched behind  him running from the cop. the pavement splashed and blood spatters as he ran. he pushed through the cops trying to evade them.  they caught him, and held him to the floor to cuffs him.  he took off running as fast as he could while demanding their id's &
I collapse in the hallway .  he punched me hard in the shoulder, hard.  i staggered back.  after a few moments, i regained my balance.  i was able to get away from him.
\\
\hline
Mom took a deep breath and stepped onto the lawn. A female lion roared. Mom dropped to her knee and lifted her head. The lion got off its lazy lazily chest. The lion was quite a lot bigger than which Mom--The lion snarled at Mom. Mom ducked and the lion howled. It charged. Better call the boys from the other room, she said and rushed to the  &
I sucked in my breath . the car hit me hard. my right hand and shoulder started to bleed. i then realized what happened. i got out of my car to check my luck. &
"No child,  he punched me in the face.   i fell to the ground with a forceful impact, my jaw broken.   when he kicked me again, my nose ended up broken.  the ambulance came into the room.  after my nose was bandaged, i was allowed to go home.
\\
\hline
Tom always wants to go where the action is.The weapons beware.He will be clutching his shield and at the ready.Place tom here as a warrior with his shield and gladius.He must end up here.Ask the computer to generate as many options as possible.Afterthinker has generated a number of attacks and defenses for tom.Tom,You see  &
It was a hand and we murdered her with her own knife.  As the cop stood on the corner, he gave chase.   as he turned the corner, there was a bullet hole.   before he knew it, the murderer fled down the street. &
I half-smiled  as he punched me hard in the nose.  my first blow didn't hurt as hard as i thought it would, but it hurt hard.   i raised my arm back up, expecting more.   instead, he swung again with all his might.  it was a swift blow to the nose.  Sam0 stared at the  ground. he raised his fist.  before anyone could move, he launched it back.  the second time it hit the ground its fist ended up broken.
\\
\hline
Mark and Frank were talking over a beer about their times together in the Marine Corps. Frank was the instigator of the conversation. "I was always the last to leave the barracks, you know. All the ones that were from the same country in the same unit. I volunteered to guard the flag, just because I was the only one in the unit that could make the others pay attention to the regulations and stop  &
Think they followed us.  they saw the cops car behind us. they came to my aid. i jumped out in front of the car!  two cops sprinted, looking terrified.  they dared me to turn around and leave. &
She was almost disappointed . she punched him again hard.  his shoulder hurt, but he tried to ignore it.  she kicked him in the face again.  he fell to the ground, unconscious.
\\
\hline
Quinton was capable of easily defeating a single orc. The orc's attacks were ineffectual. If one should get his hands on Quinton, things would become tricky, as he was well equipped, but he was not invincible. However, right now there was no orc to defeat. They were fighting a subterranean regiment, which meant that the overwhelming advantage was now on their  &
Son, I must ask  the police for assistance with his vehicle. he is driving his car onto the road when it starts to rain. the car begins to roll and is spinning off the road. son decides to wait for the rain to stop. unfortunately it never stops and son gets severely wet. &
And so, the angel  punched me in the face.  i turned around and slapped her back. she swung back with a swift kick.  before she could do another hit, i blocked it.   then, i defended myself.
\end{tabular}
\label{tab:samples}
\end{table*}










\clearpage

\section{Pseudo labels}

\citep{van2020scan} uses a contrastive learning setup similar to SimCLR \citep{Chen2020SimCLR} for pretraining an image embedder on an unlabelled image dataset. KNN is then used to cluster the embedded dataset and label new images. This method outperforms previous unsupervised classification approaches by a significant margin.

For our approach, we start by clustering embedded reviews from the Story-Critique dataset. For the embedding process, we use CARPs review embedder, followed by dimensionality reduction from $\mathbb{R}^{2048}$ to $\mathbb{R}^2$ with UMAP \citep{mcinnes2018umap}. We chose UMAP, as opposed to PCA or tSNE \citep{Maaten2008TSNE}, since CARPs latent space exists on a manifold: the unit $R^{2048}$ n-sphere. We use cosine distance as a metric since CARP was originally trained via a cosine similarity based loss. We then use hierarchical density-based clustering (HDBSCAN) \citep{mcinnes2017hdbscan} in this low dimensional space. One caveat with HDBSCAN is that it can deem some of the data as noise, leaving it unclustered. We chose $\mathbb{R}^{2}$ as a target space for UMAP as it minimized the proportion of embeddings left unclustered by HDBSCAN. The resulting clusters were found, through hand labelling, to mostly correspond with high level story features (e.g. use of dialogue, use of humor, discussion of music/instruments). To generate the centroids, we took the mean of the embeddings from the most promising clusters (with respect to hand labelling) in the original $\mathbb{R}^{2048}$ latent space, then projected it back onto the unit n-sphere. The centroids can then be used as classifiers by taking cosine similarity against them in CARPs latent space for stories (with its story embedder) or reviews (with its review embedder).  

In order to generate a pseudolabeled dataset from the (story, review) pairs in the Story-Critique dataset, we first prune pairs whose embedded review does not have sufficiently high cosine similarity with any one centroid (using a threshold of twice the average cosine similarity). We then take the index of the centroid that maximizes similarity to the review embedding, and assign this as a pseudolabel to the paired story. The sample used to generate the centroids and the sample used to generate the pseudolabeled dataset are both independently sampled subsets of the Story-Critique dataset.

Captions for the clusters, as well as more details on the hand labelling process, are available within the appendix, along with plots of the most promising clusters in the latent space. We also plan to release a version of CARP tuned on pseudo labels.

\section{COOP}


To train Psuedo CoOp for classification, we follow a similar approach to \citep{pham2021meta} in that we use CARP as a teacher to generate pseudo labels for Pseudo CoOp as a student. For Pseudo CoOp, we only use a subset of 6 centroids (see appendix) to generate pseudolabels. Additionally, the pruning described previously in the pseudo labelling section ensures a strong training signal for Pseudo CoOp. Once pseudolabels are generated and assigned to the passages, we minimize a negative log-likelihood loss.

\section{Pseudo Labelling Details}


The projected latent space and extracted hand-picked clusters can be seen in \ref{fig:carplatent}. Clusters were hand picked by sampling reviews from each HDBSCAN cluster and manually labelling them for apparent patterns. Several clusters overlapped in story feature, and several had no obvious underlying feature (i.e. they were ambiguous). To resolve this, we dropped points within ambiguous clusters, and merged nearby clusters with the same feature. Below is a table with captions for all the hand picked clusters as well as a color to match captions in the table to clusters in \ref{fig:carpclusters}. The captions for the 6 clusters whose centroids for used for training Pseudo CoOp are bolded.

\begin{center}
\begin{tabular}{c|c}
    Caption & Colour\\
    characters laughing or finding things funny & \textcolor[rgb]{0.65, 0.0, 0.15}{$\bullet$}\\
    \textbf{imagery/descriptions} & \textcolor[rgb]{0.68, 0.03, 0.15}{$\bullet$}\\
    horses & \textcolor[rgb]{0.72, 0.07, 0.15}{$\bullet$}\\
    paragraph & \textcolor[rgb]{0.75, 0.1, 0.15}{$\bullet$}\\
    characters asking questions & \textcolor[rgb]{0.79, 0.14, 0.15}{$\bullet$}\\
    dialogue as paragraph & \textcolor[rgb]{0.83, 0.18, 0.15}{$\bullet$}\\
    scene or ending & \textcolor[rgb]{0.85, 0.21, 0.16}{$\bullet$}\\
    abrupt transition & \textcolor[rgb]{0.88, 0.26, 0.19}{$\bullet$}\\
    info-dump & \textcolor[rgb]{0.9, 0.31, 0.21}{$\bullet$}\\
    connections between sentences& \textcolor[rgb]{0.92, 0.35, 0.23}{$\bullet$}\\
    merge sections or cut them & \textcolor[rgb]{0.94, 0.4, 0.25}{$\bullet$}\\
    humor & \textcolor[rgb]{0.96, 0.45, 0.27}{$\bullet$}\\
    surprise/plot twist & \textcolor[rgb]{0.97, 0.49, 0.29}{$\bullet$}\\
    single character & \textcolor[rgb]{0.97, 0.54, 0.31}{$\bullet$}\\
    swimming/sailing/ocean & \textcolor[rgb]{0.98, 0.59, 0.34}{$\bullet$}\\
    \textbf{religion} & \textcolor[rgb]{0.99, 0.64, 0.36}{$\bullet$}\\
    this is true & \textcolor[rgb]{0.99, 0.69, 0.38}{$\bullet$}\\
    long sentences & \textcolor[rgb]{0.99, 0.72, 0.42}{$\bullet$}\\
    edits to wording & \textcolor[rgb]{0.99, 0.76, 0.44}{$\bullet$}\\
    \textbf{accident/disaster} & \textcolor[rgb]{0.99, 0.79, 0.47}{$\bullet$}\\
    tired and sleepiness, dreams during sleep & \textcolor[rgb]{1.0, 0.83, 0.51}{$\bullet$}\\
    gross smells and things to see & \textcolor[rgb]{1.0, 0.87, 0.54}{$\bullet$}\\
    drinks & \textcolor[rgb]{1.0, 0.9, 0.58}{$\bullet$}\\
    food & \textcolor[rgb]{1.0, 0.92, 0.62}{$\bullet$}\\
    facial descriptions, mainly hair & \textcolor[rgb]{1.0, 0.94, 0.65}{$\bullet$}\\
    punctuation and word choice suggestions & \textcolor[rgb]{1.0, 0.96, 0.69}{$\bullet$}\\
    confusing dialogue & \textcolor[rgb]{1.0, 0.99, 0.73}{$\bullet$}\\
    clunky/confusing word usage & \textcolor[rgb]{0.99, 0.99, 0.73}{$\bullet$}\\
    internal monologues & \textcolor[rgb]{0.96, 0.98, 0.69}{$\bullet$}\\
    trains and tracks & \textcolor[rgb]{0.93, 0.97, 0.65}{$\bullet$}\\
    solar system description & \textcolor[rgb]{0.9, 0.96, 0.62}{$\bullet$}\\
    confusing word choices & \textcolor[rgb]{0.87, 0.95, 0.58}{$\bullet$}\\
    character traits & \textcolor[rgb]{0.84, 0.93, 0.54}{$\bullet$}\\
    making suggestions for changes & \textcolor[rgb]{0.8, 0.92, 0.51}{$\bullet$}\\
    cars & \textcolor[rgb]{0.76, 0.9, 0.49}{$\bullet$}\\
    feeling nervous & \textcolor[rgb]{0.73, 0.88, 0.46}{$\bullet$}\\
    scenery descriptions & \textcolor[rgb]{0.69, 0.87, 0.44}{$\bullet$}\\
    dialogue & \textcolor[rgb]{0.65, 0.85, 0.42}{$\bullet$}\\
    physical touch & \textcolor[rgb]{0.61, 0.83, 0.41}{$\bullet$}\\
    facial expressions & \textcolor[rgb]{0.56, 0.81, 0.41}{$\bullet$}\\
    dogs and cats & \textcolor[rgb]{0.51, 0.79, 0.4}{$\bullet$}\\
    war and soldiers & \textcolor[rgb]{0.46, 0.77, 0.39}{$\bullet$}\\
    school & \textcolor[rgb]{0.42, 0.75, 0.39}{$\bullet$}\\
    crimes & \textcolor[rgb]{0.36, 0.72, 0.38}{$\bullet$}\\
    guns and bombs & \textcolor[rgb]{0.31, 0.7, 0.36}{$\bullet$}\\
    \textbf{fighting} & \textcolor[rgb]{0.25, 0.67, 0.35}{$\bullet$}\\
    \textbf{music} & \textcolor[rgb]{0.19, 0.64, 0.34}{$\bullet$}\\
    \textbf{family}  & \textcolor[rgb]{0.13, 0.61, 0.32}{$\bullet$}\\
    confusing sentences & \textcolor[rgb]{0.1, 0.59, 0.31}{$\bullet$}\\
    ages of characters & \textcolor[rgb]{0.08, 0.55, 0.29}{$\bullet$}\\
    death & \textcolor[rgb]{0.06, 0.51, 0.27}{$\bullet$}\\
    murder & \textcolor[rgb]{0.04, 0.47, 0.25}{$\bullet$}\\
    relationship conflicts & \textcolor[rgb]{0.02, 0.44, 0.23}{$\bullet$}\\
    bad parents & \textcolor[rgb]{0.0, 0.41, 0.22}{$\bullet$}\\

\end{tabular}
\end{center}

\end{document}